\documentclass[runningheads]{llncs}
\usepackage{graphicx}
\usepackage{multirow}
\usepackage{subfig}
\graphicspath{ {./pictures/} }

\begin{document}

\title{Predicting Medical Interventions from Vital
Parameters: Towards a Decision Support System
for Remote Patient Monitoring}

\titlerunning{Predicting Medical Interventions from Vital
Parameters}

\author{Kordian~Gontarska\inst{1,2,3}\orcidID{0000-0002-2755-405X} \and
Weronika~Wrazen\inst{1} \and
Jossekin~Beilharz\inst{1,3} \and
Robert~Schmid\inst{1,3} \and
Lauritz~Thamsen\inst{2} \and
Andreas~Polze\inst{1}}

\authorrunning{Gontarska, K. et al.}

\institute{Hasso Plattner Institute, University of Potsdam, Germany,
\email{firstname.lastname@hpi.de} \and
Technische Universitat Berlin, Germany,
\email{lauritz.thamsen@tu-berlin.de} \and
Charité – Universitätsmedizin Berlin, corporate member of Freie Universität Berlin,
Humboldt-Universität zu Berlin, and Berlin Institute of Health}

\maketitle

\begin{abstract}
Cardiovascular diseases and heart failures in particular are the main cause of non-communicable disease mortality in the world. 
Constant patient monitoring enables better medical treatment as it allows practitioners to react on time and provide the appropriate treatment. 
Telemedicine can provide constant remote monitoring so patients can stay in their homes, only requiring medical sensing equipment and network connections. 
A limiting factor for telemedical centers is the amount of patients that can be monitored simultaneously. 
We aim to increase this amount by implementing a decision support system. This paper investigates a machine learning model to estimate a risk score based on patient vital parameters that allows sorting all cases every day to help practitioners focus their limited capacities on the most severe cases.
The model we propose reaches an AUCROC of 0.84, whereas the baseline rule-based model reaches an AUCROC of 0.73. Our results indicate that the usage of deep learning to improve the efficiency of telemedical centers is feasible. This way more patients could benefit from better health-care through remote monitoring.

\end{abstract}

\section{Introduction} \par
According to the World Health Organization, cardiovascular diseases (CVDs) are the main cause of a non-communicable disease mortality in the world~\cite{WHO_2020}. 
It is important to detect a patient's critical condition early to enable a timely intervention. One way to ensure this is to monitor patients remotely in their homes from telemedical centers (TMCs). Modern technology makes it possible to provide patients with monitoring services even in areas without comprehensive medical infrastructures. In recent years, it was shown that telemedical interventions reduce the mortality in patients with heart failures~\cite{koehler-timhf2,telemedicine_effectiveness}. 
\par This paper is a part of the Telemed5000 project and follows our previous work on clinical decision support systems for heart failure which was a part of the Fontane project in collaboration with Charité, Berlin~\cite{fontane,heinze_hybrid_remote_monitoring}. Our aim is to scale up the TMC capacity to ensure that up to 5,000 patients will be cared for in the future utilizing Artificial Intelligence (AI).
\par In this paper we describe the development and evaluation of a machine learning model for the prediction of the daily per-patient risk of being in a medically critical condition.
The patients are sorted by this estimated risk, enabling the TMC to concentrate on the most severe cases.
To accomplish this we use a database with daily vital parameters from the TIM-HF 2 study~\cite{koehler-timhf2}.
 
\section{Materials and Methods}

In this research we used the Telemedical Interventional Management in Heart Failure II (TIM-HF2) database, which was created by Charité, Berlin during the Fontane project~\cite{koehler-timhf2}. The trial has been conducted in Germany between 2013 and 2018. TIM-HF2 included 1,538 patients (773 usual care) whose stage of heart failure is classed II or III according to the New York Health Association (NYHA) classification. Additionally they were admitted to a hospital at most 12 months prior to the study due to heart failure and had a left ventricular ejection fraction (LVEF) of $<$ 45\%.
The dataset includes daily measurements performed by the patients themselves using a weight scale, a blood pressure monitor, a pulse oximeter, a small ECG device and a tablet-like device for the self-reporting of their well-being. In total the unprocessed dataset consists of records from 763 patients out of which 100 died before the end of the study (66 within 7 days after their last measurement). The database also contains clinical events like 387 endpoint-adjudicated hospitalizations and 4,329 interventions performed by the TMC. Patients were asked to participate for one year.

We included the following features into the data: age, weight, blood pressure, oxygen saturation, gender, diabetes, the NYHA class, several symptoms and signs of heart failure (e.g. AV Block, LBBB), automatically extracted data from ECG (heart rate, sinus rhythm, ventricular tachycardia, atrial fibrillation), self-assessed state of health, weight difference (1, 3 and 8 days difference), social variables (e.g. living alone, anxiety). The binary predictor variable is a union of TMC's intervention, hospitalization or death events. 
Missing values were linearly imputed for up to 2 consecutive missing days, the rest got dropped. The positive class forms only approximately 2\% of the dataset, thus we oversampled observations from the minority class in the training set to balance the classes.
The dataset was split into three sets: train, validation, and test in a proportion 4:1:1 respectively. The distribution of samples and events per patient was preserved across all sets. Each patient was assigned to exactly one set. To evaluate model performance we took the following metrics into consideration: Receiver Operating Characteristic (ROC) curve, area under ROC curve (AUCROC), Precision - Recall curve, and area under PR curve (AUCPR).

We investigate different deep neural network (DNN) models and compare them to a rule-based baseline. The rule-based model (RB) is based off the TIM-HF 2 study~\cite{koehler-timhf2}. The rules are heart related and consist of engineered features and thresholds defined by an expert group at the Charité \cite{koehler-studydesign}. All DNN models had an output layer with \textit{a sigmoid activation} function, \textit{binary cross-entropy} as a loss function, and \textit{Adam} as the optimization algorithm. We tested between 2 and 5 hidden layers with 5 to 150 neurons in each. Additionally we tested  linear, sigmoid, and ReLU activation functions, and dropout rates between 0 and 0.5.

\section{Results}
\begin{figure}[h!]
	\centering
	\hspace{\fill}%
	\subfloat[ROC Curves\label{fig:roc_curve}]{
		\includegraphics[width=0.4\textwidth]{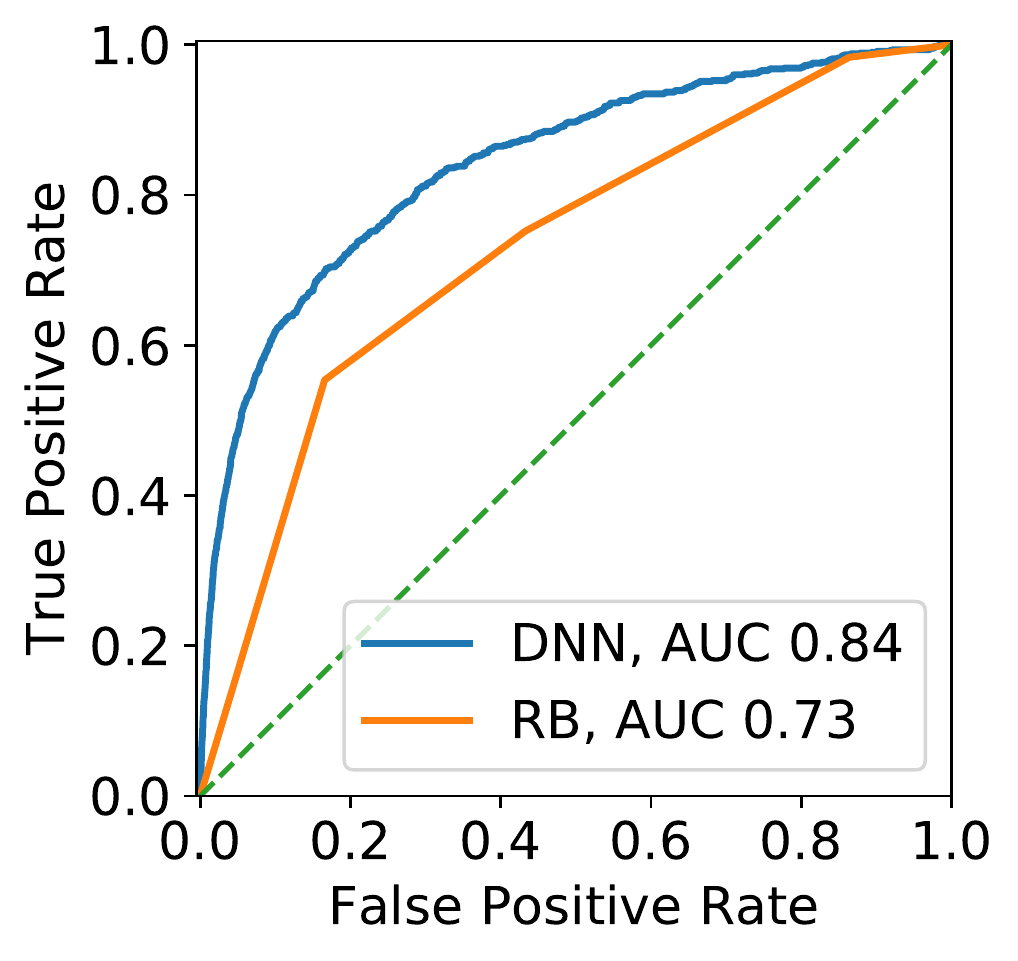}
	}
	\hspace{\fill}%
	\subfloat[PR Curves\label{fig:pr_curve}]{
		\includegraphics[width=0.4\textwidth]{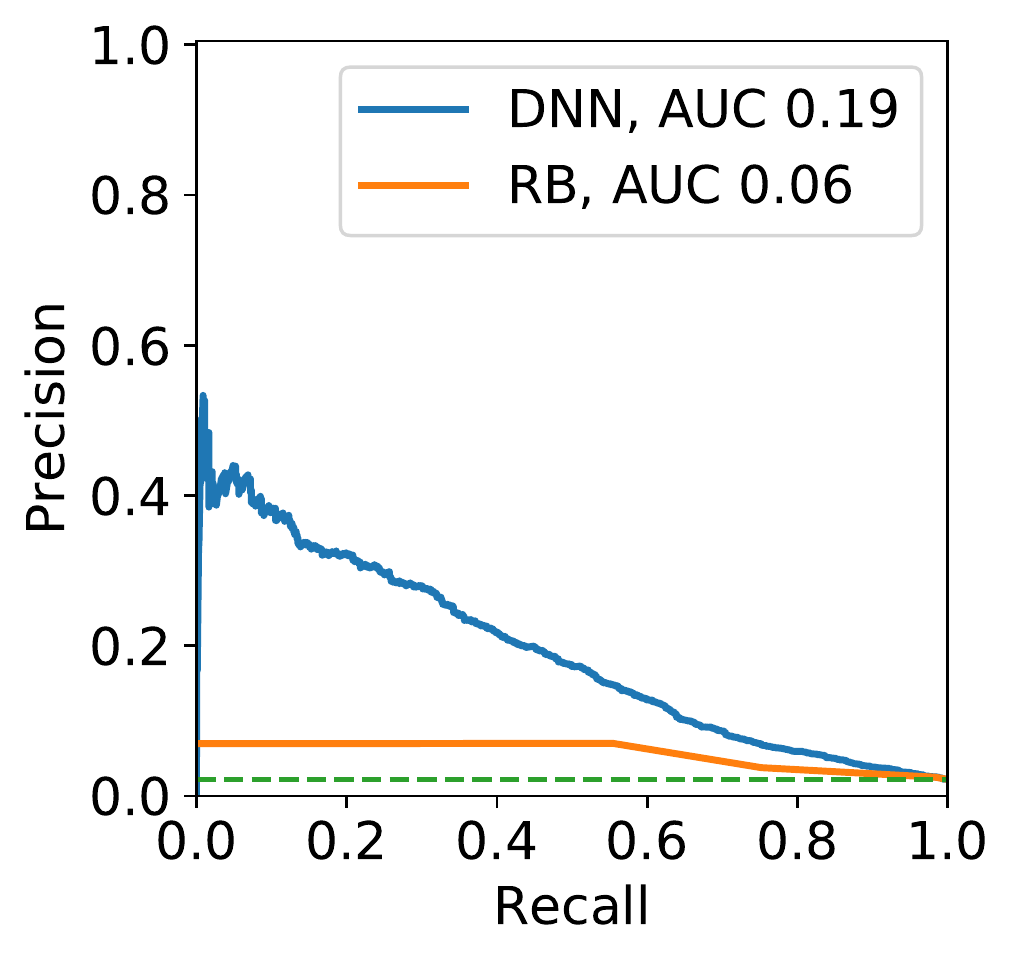}
	}
	\hspace*{\fill}%
	\caption{The figures show the (a) ROC curves and (b) PR curves for both the DNN and the rule-based model. The dashed lines represent what performance a purely random classifier would achieve.} \label{roc_pr_curve}
\end{figure}

\begin{figure}[h!]
	\centering
	\hspace{\fill}%
	\subfloat[Deep Neural Network]{
		\includegraphics[width=0.45\textwidth]{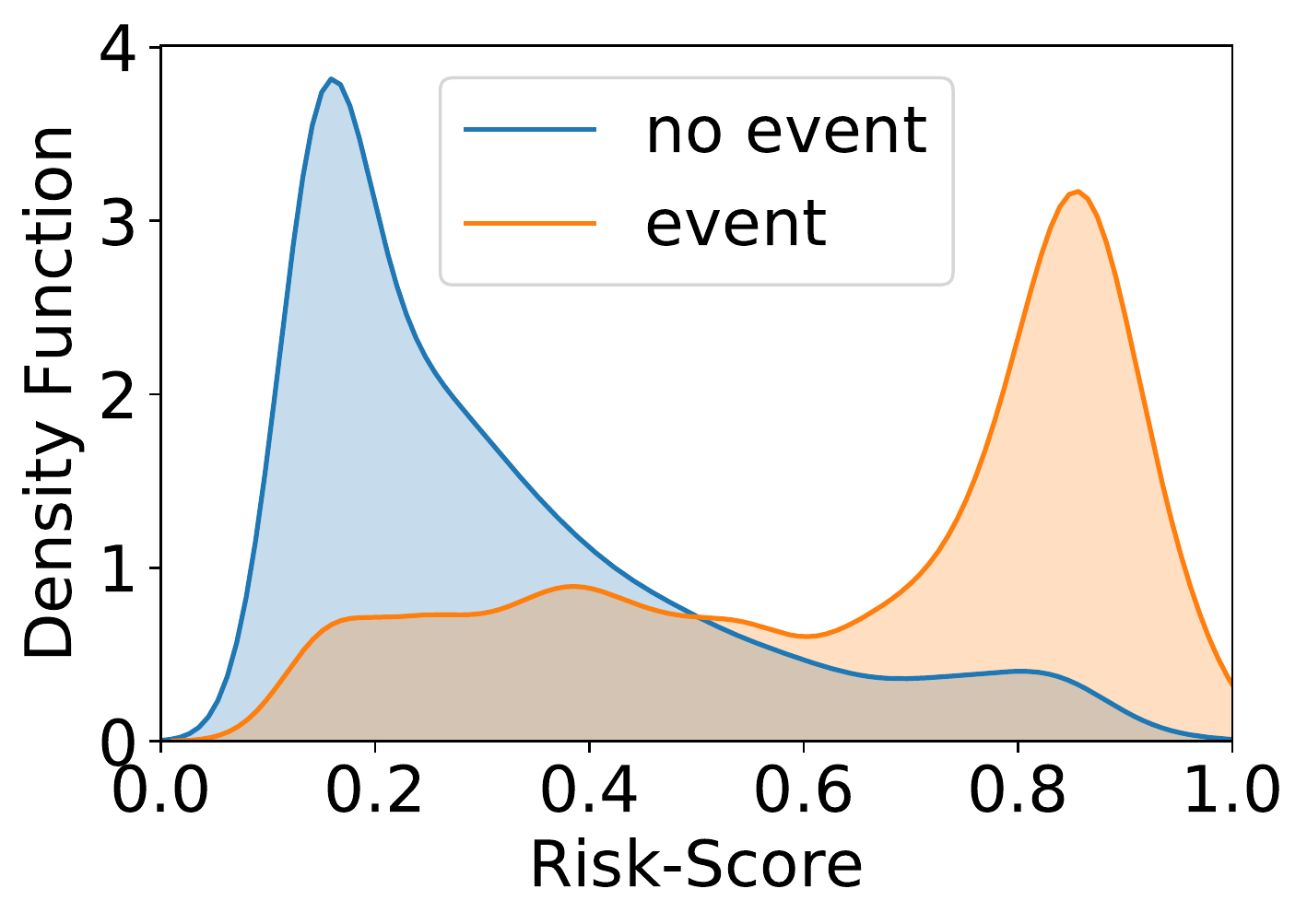}
	}
	\hspace{\fill}%
	\subfloat[Rule-Based Model]{
		\includegraphics[width=0.47\textwidth]{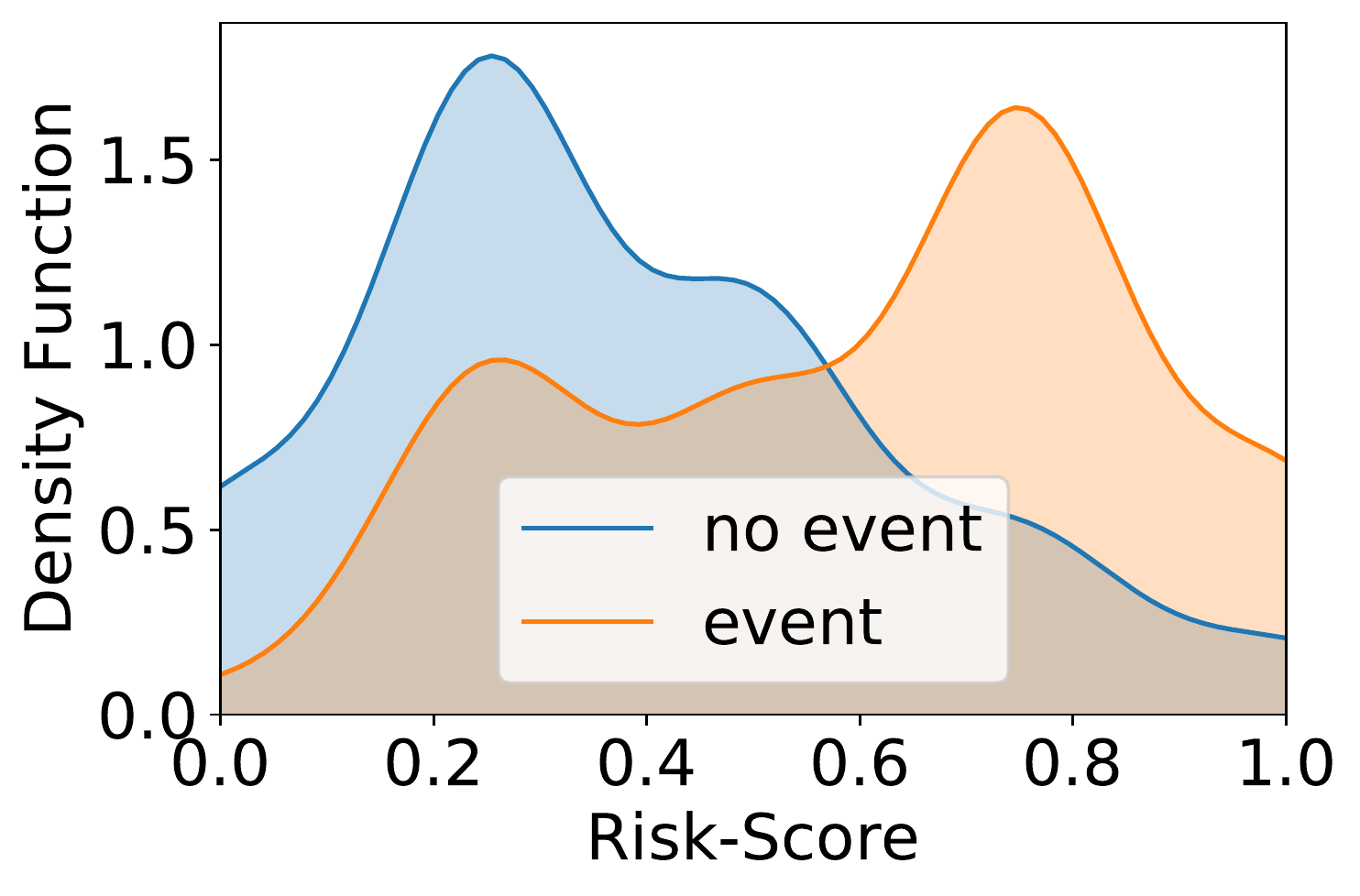}
	}
	\hspace*{\fill}%
	\caption{The depicted plots show the distribution of the predicted risk-scores in the test set, separated by the true label.} \label{prob_distr}
\end{figure}

Fig. \ref{fig:roc_curve} shows the ROC curves for the selected DNN and the rule-based model. The DNN outperforms the rule-based model having better sensitivity at any specificity and an AUCROC of 0.84 as compared to 0.73. Fig. \ref{fig:pr_curve} shows the Precision/Recall curve for both models. The DNN outperforms the rule-based model in precision at any recall rates. The plots in Fig. \ref{prob_distr} show the distributions of the predicted risk-scores
for both classes, as predicted by the DNN and rule-based models. It can be seen that the DNN model performs better in detecting events than the rule-based model, as there is a clearer cut between the distributions. The final DNN model was trained for 453 epochs using a batch size of 4096, and a learning rate of 0.001. It has 3 hidden layers with 35, 20, and 35 neurons respectively. All neurons in the hidden layers use ReLU as their activation function and have dropout rates of 0.25, 0.15, and 0.3. The patient's self assessment, weight differences, pulse-rate, and complaints had the highest impact on the models decision making.

\section{Related Work} \par
Decision Support Systems (DSSs) have been used in the medical field since the seventies~\cite{shortliffe1975computer}. Seto et al. applied a rule-based model to monitor patients with heart failure~\cite{rulebased_telemedicine_hf}. A rule-based model was implemented for the Fontane project, which had to prioritize patients based on their daily vital parameters~\cite{koehler-timhf2,fontane}.
Groccia et. al. proposed a linear Support Vector Machine (SVM) model which predicts major cardiovascular worsening events for patients with heart failure~\cite{worsening_hf_ml}. 
Stehlik et. al. studied the potential efficiency of noninvasive remote monitoring in predicting heart failure re-hospitalizations ~\cite{stehlik_hf_hosp_pred}. 
Heinze et al. proposed a Hybrid AI model as an improvement for the rule-based model in Fontane~\cite{heinze_hybrid_remote_monitoring}. The hybrid model consists of a Neural Network (NN) with one hidden layer and two rules which were handcrafted by medical experts. 
\section{Conclusion}
Based on the dataset of daily recordings of vital parameters, medical interventions, hospitalizations and deaths we developed a machine learning model to predict the risk of a patient requiring an intervention. 
We showed that our approach outperforms the rule-based model used in the Fontane project. The DNN may help a medical practitioner to provide valuable assessment to more critical patients on time.
To ensure that no patient is overseen, investigations in the TMC could be scheduled in addition to the model's sorting. This would reduce the capacity for the model but ensures that each patient is going to be seen within a defined time period (e.g. 14 days).
In further research we will investigate Recurrent Neural Networks (RNN), because of the time series nature of the dataset. The model devised in this research is not patient specified but generalized among all patients. It can be assumed that there is some variance between the patients which could be used to adapt the model to each patient individually and thus boosting its performance. 
\section{Acknowledgment} \par
We thank Prof. Dr. med. Friedrich Köhler and his team for the provisioning of the dataset and Prof. Dr. med. Alexander Meyer for helping and supporting us in analyzing the dataset. We are grateful to Alexander Acker for the fruitful discussions, Volker Möller for his help with the dataset, and Boris Pfahringer for his valuable feedback on our evaluation. This research and the Telemed5000 project have been supported by the Federal Ministry for Economic Affairs and Energy of Germany as part of the program "Smart Data" (project
number 01MD19014C). 

\bibliography{references}

\begin{thebibliography}{10}
\providecommand{\url}[1]{\texttt{#1}}
\providecommand{\urlprefix}{URL }
\providecommand{\doi}[1]{https://doi.org/#1}

\bibitem{telemedicine_effectiveness}
Ekeland, A., Bowes, A., Flottorp, S.: Effectiveness of telemedicine: A
  systematic review of reviews. International Journal of Medical Informatics
  \textbf{79} (2010)

\bibitem{worsening_hf_ml}
Groccia, M., Lofaro, D., Guido, R., Conforti, D., Sciacqua, A.: Predictive
  models for risk assessment of worsening events in chronic heart failure
  patients. CinC'18 (2018)

\bibitem{heinze_hybrid_remote_monitoring}
Heinze, T., Wierschke, R., Schacht, A., Löwis, M.: A hybrid artificial
  intelligence system for assistance in remote monitoring of heart patients.
  HAIS'11, vol.~6679 (2011)

\bibitem{koehler-timhf2}
Koehler, F., Koehler, K., Deckwart, O., Prescher, S., Wegscheider, K., Kirwan,
  B.A., Winkler, S., Vettorazzi, E., Bruch, L., Oeff, M., Zugck, C., Doerr, G.,
  Naegele, H., Butter, C., Sechtem, U., Angermann, C., Gola, G., Prondzinsky,
  R., Stangl, K.: Efficacy of telemedical interventional management in patients
  with heart failure (tim-hf2): a randomised, controlled, parallel-group,
  unmasked trial. The Lancet  (2018)

\bibitem{koehler-studydesign}
Koehler, F., Koehler, K., Deckwart, O., Prescher, S., Wegscheider, K., Winkler,
  S., Vettorazzi, E., Polze, A., Stangl, K., Hartmann, O., Marx, A., Neuhaus,
  P., Scherf, M., Kirwan, B.A., Anker, S.: Telemedical interventional
  management in heart failure ii (tim-hf2), a randomised, controlled trial
  investigating the impact of telemedicine on unplanned cardiovascular
  hospitalisations and mortality in heart failure patients: study design and
  description: Tim-hf2: study design. European Journal of Heart Failure
  \textbf{20} (2018)

\bibitem{fontane}
Polze, A., Tröger, P., Hentschel, U., Heinze, T.: A scalable, self-adaptive
  architecture for remote patient monitoring. ISORCW'10 (2010)

\bibitem{rulebased_telemedicine_hf}
Seto, E., Leonard, K., Cafazzo, J., Barnsley, J., Masino, C., Ross, H.:
  Developing healthcare rule-based expert systems: Case study of a heart
  failure telemonitoring system. International Journal of Medical Informatics
  \textbf{81} (2012)

\bibitem{shortliffe1975computer}
Shortliffe, E.H., Davis, R., Axline, S.G., Buchanan, B.G., Green, C.C., Cohen,
  S.N.: Computer-based consultations in clinical therapeutics: explanation and
  rule acquisition capabilities of the mycin system. Computers and Biomedical
  Research  \textbf{8} (1975)

\bibitem{stehlik_hf_hosp_pred}
Stehlik, J., Schmalfuss, C., Bozkurt, B., Nativi-Nicolau, J., Wohlfahrt, P.,
  Wegerich, S., Rose, K., Ray, R., Schofield, R., Deswal, A., Sekaric, J.,
  Anand, S., Richards, D., Hanson, H., Pipke, M., Pham, M.: Continuous wearable
  monitoring analytics predict heart failure hospitalization: The link-hf
  multicenter study. Circulation: Heart Failure  \textbf{13} (2020)

\bibitem{WHO_2020}
WHO: World health statistics 2020: monitoring health for the sdgs, sustainable
  development goals  (2020)

\end{thebibliography}
\bibliographystyle{splncs04}

\end{document}